\documentclass[english,letterpaper,sort&compress]{article}
\usepackage[T1]{fontenc}
\usepackage[latin9]{inputenc}
\usepackage{amsmath}
\usepackage{amssymb}
\usepackage{graphicx}
\usepackage[numbers]{natbib}

\makeatletter
 \usepackage[final]{nips_2016}
 \usepackage{times}
\usepackage{algorithm,algpseudocode}

\makeatother

\usepackage{babel}
\begin{document}

\title{Human Decision-Making under Limited Time}

\author{
  Pedro A. Ortega\\
  Department of Psychology\\
  University of Pennsylvania\\
  Philadelphia, PA 19104\\
  \texttt{ope@seas.upenn.edu}
  \And
  Alan A. Stocker\\
  Department of Psychology\\
  University of Pennsylvania\\
  Philadelphia, PA 19014\\
  \texttt{astocker@sas.upenn.edu}
}

\maketitle
\begin{abstract}
Subjective expected utility theory assumes that decision-makers possess
unlimited computational resources to reason about their choices; however,
virtually all decisions in everyday life are made under resource constraints---i.e.~decision-makers
are bounded in their rationality. Here we experimentally tested the
predictions made by a formalization of bounded rationality based on
ideas from statistical mechanics and information-theory. We systematically
tested human subjects in their ability to solve combinatorial puzzles
under different time limitations. We found that our bounded-rational
model accounts well for the data. The decomposition of the fitted
model parameter into the subjects' expected utility function and resource
parameter provide interesting insight into the subjects' information
capacity limits. Our results confirm that humans gradually fall back
on their learned prior choice patterns when confronted with increasing
resource limitations.
\end{abstract}

\section{Introduction}

Human decision-making is not perfectly rational. Most of our choices
are constrained by many factors such as perceptual ambiguity, time,
lack of knowledge, or computational effort \citep{Gigerenzer2001}.
Classical theories of rational choice do not apply in such cases because
they ignore information-processing resources, assuming that decision-makers
always pick the optimal choice \citep{Rubinstein1998}. However, it
is well known that human choice patterns deviate qualitatively from
the perfectly rational ideal with increasing resource limitations.

It has been suggested that such limitations in decision-making can
be formalized using ideas from statistical mechanics \citep{Ortega2013}
and information theory \citep{Tishby2011}. These frameworks propose
that decision-makers act \emph{as if} their choice probabilities were
an optimal compromise between maximizing the expected utility and
minimizing the KL-divergence from a set of prior choice probabilities,
where the trade-off is determined by the amount of available resources.
This optimization scheme reduces the decision-making problem to the
\emph{inference} of the optimal choice from a stimulus, where the
likelihood function results from a combination of the decision-maker's
subjective preferences and the resource limitations.

The aim of this paper is to systematically validate the model of bounded-rational
decision-making on human choice data. We conducted an experiment in
which subjects had to solve a sequence of combinatorial puzzles under
time pressure. By manipulating the allotted time for solving each
puzzle, we were able to record choice data under different resource
conditions. We then fit the bounded-rational choice model to the dataset,
obtaining a decomposition of the choice probabilities in terms of
a resource parameter and a set of stimulus-dependent utility functions.
Our results show that the model captures very well the gradual shifts
due to increasing time constraints that are present in the subjects'
empirical choice patterns. 

\newpage{}

\section{A Probabilistic Model of Bounded-Rational Choices\label{sec:model}}

We model a bounded-rational decision maker as an expected utility
maximizer that is subject to information constraints. Formally, let $\mathcal{X}$
and $\mathcal{Y}$ be two finite sets, the former corresponding to
a \emph{set of stimuli} and the latter to a \emph{set of choices}; and
let $P(y)$ be a prior distribution over optimal choices $y\in\mathcal{Y}$
that the decision-maker may have learned from experience. When presented
with a stimulus $x\in\mathcal{X}$, a bounded-rational decision-maker
transforms the prior choice probabilities $P(y)$ into posterior choice
probabilities $P(y|x)$ and then generates a choice according to $P(y|x)$.

This transformation is modeled as the optimization of a regularized
expected utility known as the \emph{free energy functional}:

\begin{equation}
F\bigl[Q(y|x)\bigr]:=\underbrace{\sum_{y}Q(y|x)U_{x}(y)}_{\text{Expected Utility}}-\underbrace{\frac{1}{\beta}\sum_{y}Q(y|x)\log\frac{Q(y|x)}{P(y)}}_{\text{Regularization}},\label{eq:fe}
\end{equation}
where the posterior is defined as the maximizer $P(y|x):=\arg\max_{Q(y|x)}F[Q(y|x)]$.
Crucially, the optimization is determined by two factors. The first
is the decision-maker's \emph{subjective utility function}~$U_{x}:\mathcal{Y}\rightarrow\mathbb{R}$
encoding the desirability of a choice~$y$ given a stimulus~$x$.
The second is the \emph{inverse temperature}~$\beta$, which determines
the resources of deliberation available for the decision-task%
\footnote{For simplicity, here we consider only strictly positive values for
the inverse temperature $\beta$, but its domain can be extended to
negative values to model other effects, e.g.~risk-sensitive estimation
\citep{Ortega2013}.%
}, but which are neither known to, nor controllable by the decision-maker.
The resulting posterior has an analytical expression given by the
Gibbs distribution

\begin{equation}
P(y|x)=\frac{1}{Z_{\beta}(x)}P(y)\exp\bigl\{\beta U_{x}(y)\bigr\},\label{eq:posterior}
\end{equation}
where $Z_{\beta}(x)$ is a normalizing constant \citep{Ortega2013}.
The expression~(\ref{eq:posterior}) highlights a connection to inference:
bounded-rational decisions can also be computed via Bayes' rule in
which the likelihood is determined by $\beta$ and $U_{x}$ as follows:

\begin{equation}
P(y|x)=\frac{P(y)P(x|y)}{\sum_{y'}P(y')P(x|y')},\quad\text{hence}\quad P(x|y)\propto\exp\bigl\{\beta U_{x}(y)\bigr\}.\label{eq:likelihood}
\end{equation}
The objective function (\ref{eq:fe}) can be motivated as a trade-off
between maximizing expected utility and minimizing information cost~\citep{Tishby2011,Ortega2013}.
Near-zero values of $\beta$, which correspond to heavily-regularized
decisions, yield posterior choice probabilities that are similar to
the prior. Conversely, with growing values of~$\beta$, the posterior
choice probabilities approach the perfectly-rational limit.

\paragraph{Connection to regret. }

Bounded-rational decision-making is related to \emph{regret theory}
\citep{Loomes1982,Fishburn1982,Bell1982}. To see this, define the
\emph{certainty-equivalent} as the maximum attainable value for (\ref{eq:fe}):
\begin{equation}
U_{x}^{\ast}:=\max_{Q(y|x)}\Bigl\{ F\bigl[Q(y|x)\bigr]\Bigr\}=\frac{1}{\beta}\log Z_{\beta}(x).\label{eq:ce}
\end{equation}
The certainty-equivalent quantifies the \emph{net worth} of the stimulus
$x$ prior to making a choice. The decision process treats (\ref{eq:ce})
as a reference utility used in the assessment of the alternatives.
Specifically, the modulation of any choice is obtained by measuring
up the utility against the certainty-equivalent: 

\begin{equation}
\underbrace{\log\frac{P(y|x)}{P(y)}}_{\text{Change of \ensuremath{y}}}=-\beta\Bigl[\underbrace{U_{x}^{\ast}-U_{x}(y)}_{\text{Regret of \ensuremath{y}}}\Bigr].\label{eq:rejoice}
\end{equation}
Accordingly, the difference in log-probability is proportional to
the negative regret \citep{Bleichrodt2015}. The decision-maker's
utility function specifies a direction of change relative to the certainty-equivalent,
whereas the strength of the modulation is determined by the inverse
temperature.

\section{Experimental Methods}

\begin{figure}[h]
\begin{centering}
\includegraphics[width=1\textwidth]{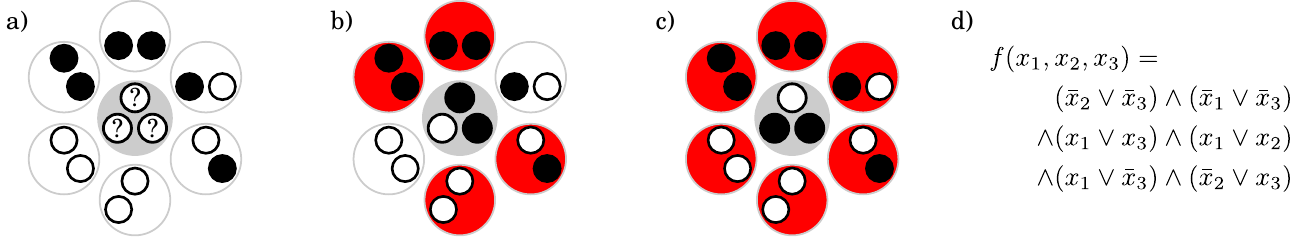}
\par\end{centering}

\caption{\emph{Example puzzle.} a) Each puzzle is a set of six circularly arranged
patches containing patterns of black ($\bullet$) and white circles
($\circ$). In each trial, the positions of the patches were randomly
assigned to one of the six possible locations. Subjects had to choose
the three center colors such that there was at least one (color and
position) match for each patch. For instance, the choice in (b) only
matches four out of six patches (in red), while (c) solves the puzzle.
The puzzle is a visualization of the Boolean formula in~(d). \label{fig:cnf-formulas}}
\end{figure}

We conducted a choice experiment where subjects had to solve puzzles
under time pressure. Each puzzle consisted of Boolean formula in conjunctive
normal form (CNF) that was disguised as an arrangement of circular
patterns (see Fig.~\ref{fig:cnf-formulas}). The task was to find
a truth assignment that satisfied the formula. Subjects could pick
an assignment by setting the colors of a central pattern highlighted
in gray. Formally, the puzzles and the assignments corresponded to
the stimuli $x\in\mathcal{X}$ and the choices $y\in\mathcal{Y}$
respectively, and the duration of the puzzle was the resource parameter
that we controlled (see equation~\ref{eq:fe}). 

We restricted our puzzles to a set of five CNF formulas having 6~clauses,
2~literals per clause, and 3~variables. Subjects were trained only
on the first four puzzles, whereas the last one was used as a control
puzzle during the test phase. All the chosen puzzles had a single solution out of the $2^{3}=8$
possible assignments. 

We chose CNF formulas because they provide a general%
\footnote{More precisely, the 2-SAT and SAT problems are NL- and NP-complete
respectively. This means that every other decision problem within
the same complexity class can be reduced (i.e.~rephrased) as a SAT
problem.%
} and flexible platform for testing decision-making behavior. Crucially,
unlike in an estimation task, finding the relation between a stimulus
and a choice is non-trivial and requires solving a computational problem.

\subsection{Data Collection}

Two symmetric versions of the experiment were conducted on Amazon
Mechanical Turk. For each, we collected choice data from 15 anonymized
participants living in the United States, totaling 30 subjects. Subjects
were paid 10~dollars for completing the experiment.  The typical
runtime of the experiment ranged between 50 and 130 minutes.

For each subject, we recorded a sequence of 90 training and 285 test
trials. The puzzles were displayed throughout the whole trial, during
which the subjects could modify their choice at will. The training
trials allowed subjects to familiarize themselves with the task and
the stimuli, whereas the test trials measured their adapted choice
behavior as a function of the stimulus and the task duration. Training
trials were presented in blocks of 18 for a long, fixed duration;
the test trials, which were of variable duration, were presented in
blocks of 19 (18 regular + 1 control trial). To avoid the collection
of poor quality data, subjects had to repeat a block if they failed
more than 6 trials within the same block, thereby setting a performance
threshold that was well above chance level. Participants could initiate
a block whenever they felt ready to proceed. Within a block, the inter-trial
durations were drawn uniformly between 0.5 and 1.5s. 

Each trial consisted of one puzzle that had to be solved within a
limited time. Training trials lasted 10s each, while test trials had
durations of 1.25, 2.5, and 5s. Apart from a visual cue shown 1s before
the end of each trial, there was no explicit feedback communicating
the trial length. Therefore, subjects did not know the duration of
individual test trials beforehand and thus \emph{could not use this
information in their solution strategy}.  A trial was considered
successful only if all the clauses of the puzzle were satisfied.

\newpage{}

\section{Analysis}

The recorded data $\mathcal{D}$ consists of a set of tuples $(x,r,y)$,
where $x\in\mathcal{X}$ is a stimulus, $r\in\mathcal{R}$ is a resource
parameter (i.e.~duration), and $y\in\mathcal{Y}$ a choice. In order
to analyze the data, we made the following assumptions:
\begin{enumerate}
\item \emph{Transient regime}: During the training trials, the subjects
converged to a set of subjective preferences over the choices which
depended only on the stimuli. 
\item \emph{Permanent regime}: During the test trials, subjects did not
significantly change the preferences that they learned during the
training trials. Specifically, choices in the same stimulus-duration
group were i.i.d.~throughout the test phase.
\item \emph{Negligible noise}: We assumed that the operation of the input
device and the cue signaling the imminent end of the trial did not
have a significant impact on the distribution over choices. 
\end{enumerate}
Our analysis only focused only the test trials. Let $P(x,r,y)$ denote
the empirical probabilities\footnote{More precisely, $P(x,r,y) \propto N(x,r,y) + 1$, 
where $N(x,r,y)$ is the count of ocurrences of $(x,r,y)$.} 
of the tuples $(x,r,y)$ estimated from
the data. From these, we derived the probability distribution $P(x,r)$
over the stimulus-resource context, the prior $P(y)$ over choices,
and the posterior $P(y|x,r)$ over choices given the context through
marginalization and conditioning.

\subsection{Inferring Preferences}

By fitting the model, we decomposed the choice probabilities into:
(a)~an inverse temperature function $\beta:\mathcal{R}\rightarrow\mathbb{R}$;
and (b)~a set of subjective utility functions $U_{x}:\mathcal{Y}\rightarrow\mathbb{R}$,
one for each stimulus~$x$. We assumed that the sets $\mathcal{X}$,
$\mathcal{R}$, and $\mathcal{Y}$ were finite, and we used vector
representations for~$\beta$ and the~$U_{x}$. To perform the decomposition,
we minimized the average Kullback-Leibler divergence
\begin{equation}
J=\sum_{x,r}P(x,r)\biggl[\sum_{y}P(y|x,r)\log\frac{P(y|x,r)}{Q(y|x,r)}\biggr],\label{eq:avg-kl}
\end{equation}
w.r.t.~the inverse temperatures $\beta(r)$ and the utilities $U_{x}(y)$
through the probabilities $Q(y|x,r)$ of the choice $y$ given the
context $(x,r)$ as derived from the Gibbs distribution
\begin{equation}
Q(y|x,r)=\frac{1}{Z_{\beta}}P(y)\exp\Bigl\{\beta(r)U_{x}(y)\Bigr\},\label{eq:model-posterior}
\end{equation}
where $Z_{\beta}$ is the normalizing constant. We used the objective
function~(\ref{eq:avg-kl}) because it is the Bregman divergence
over the simplex of choice probabilities \citep{Banerjee2005}. Thus,
by minimizing the objective function~(\ref{eq:avg-kl}) we were seeking
a decomposition such that the Shannon information contents of $P(y|x,r)$
and $Q(y|x,r)$ were matched against each other in expectation.

We minimized~(\ref{eq:avg-kl}) using gradient descent. For this,
we first rewrote~(\ref{eq:avg-kl}) as

\[
J=\sum_{x,\beta,y}P(x,r,y)\biggl\{\log\frac{P(y|x,r)}{P(y)}-\beta(r)U_{x}(y)+\log Z_{\beta}\biggr\}
\]
to expose the coordinates of the exponential manifold and then calculated
the gradient. The partial derivatives of $J$ w.r.t.~$\beta(r)$
and $U_{x}(y)$ are equal to 
\begin{align}
\frac{\partial J}{\partial\beta(r)} & =\sum_{x,y}P(x,r)\sum_{y}\Bigl[Q(y|x,r)-P(y|x,r)\Bigr]U_{x}(y)\label{eq:dJdbeta}\\
\text{and}\quad\frac{\partial J}{\partial U_{x}(y)} & =\sum_{x,y}P(x,r)\Bigl[Q(y|x,r)-P(y|x,r)\Bigr]\beta(r)\label{eq:dJdU}
\end{align}
respectively. The Gibbs distribution (\ref{eq:model-posterior}) admits
an infinite number of decompositions, and therefore we had to fix
the scaling factor and the offset to obtain a unique solution. The
scale was set by clamping the value of $\beta(r_{0})=\beta_{0}$ for
an arbitrarily chosen resource parameter $r_{0}\in\mathcal{R}$; we
used $\beta(r_{0})=1$ for $r_{0}=1$s. The offset was fixed by
normalizing the utilities. A simple way to achieve this is by subtracting
the certainty-equivalent from the utilities, i.e.~for all $(x,y)$,
\begin{equation}
U_{x}(y)\leftarrow U_{x}(y)-\frac{1}{\beta(r_{0})}\log\sum_{y}P(y)\exp\Bigl\{\beta(r_{0})U_{x}(y)\Bigr\}.\label{eq:normalize-utilities}
\end{equation}
Utilities normalized in this way are proportional to the negative
regret (see Section~\ref{sec:model}) and thus have an intuitive
interpretation as modulators of change of the choice distribution.

The resulting decomposition algorithm repeats the following two steps
until convergence: first it updates the inverse temperature and utility
functions using gradient descent, i.e. 

\begin{equation}
\beta(r)\longleftarrow\beta(r)-\eta_{t}\frac{\partial J}{\partial\beta(r)}\quad\text{and}\quad U_{x}(y)\longleftarrow U_{x}(y)-\eta_{t}\frac{\partial J}{\partial U_{x}(y)}\label{eq:sgd}
\end{equation}
for all $(r,x,y)\in\mathcal{R\times\mathcal{X}\times\mathcal{Y}}$
; and seconds it projects the parameters back onto a standard submanifold
by setting $r=r_{0}$ and normalizing the utilities in each iteration
using (\ref{eq:normalize-utilities}). For the learning rate $\eta_{t}>0$,
we choose a simple schedule that satisfied the Robbins-Monro conditions
$\sum_{t}\eta_{t}=\infty$ and $\sum_{t}\eta_{t}^{2}<\infty$.

\subsection{Expected Utility and Decision Bandwidth}

The inferred model is useful for investigating the decision-maker's
performance under different settings of the resource parameter---in
particular, to determine the asymptotic performance limits. Two quantities
are of special interest: the \emph{expected utility} averaged over
the stimuli and the \emph{mutual information} between the stimulus
and the choice, both as functions of the inverse temperature $\beta$.
Given $\beta$, we define these quantities as

\begin{equation}
EU_{\beta}:=\sum_{x,y}P(x)Q_{\beta}(y|x)U_{x}(y)\quad\text{and}\quad I_{\beta}:=\sum_{x,y}P(x)Q_{\beta}(y|x)\log\frac{Q_{\beta}(y|x)}{Q_{\beta}(y)}\label{eq:eu-i}
\end{equation}
respectively. Both definitions are based on the joint distribution
$P(x)Q_{\beta}(y|x)$ in which $Q_{\beta}(y|x)\propto P(y)\exp\{\beta U_{x}(x)\}$
is the Gibbs distribution derived from the prior $P(y)$ and the utility
functions $U_{x}(y)$. The marginal over choices is given by $Q_{\beta}(y)=\sum_{x}P(x)Q_{\beta}(y|x)$.
The mutual information~$I_{\beta}$ is a measure of the decision
bandwidth, because it quantifies the average amount of information
that the subject has to extract from the stimulus in order to produce
the choice.

\section{Results}

\subsection{Decomposition into prior, utility, and inverse temperature}

\begin{figure}
\begin{centering}
\includegraphics[width=1\textwidth]{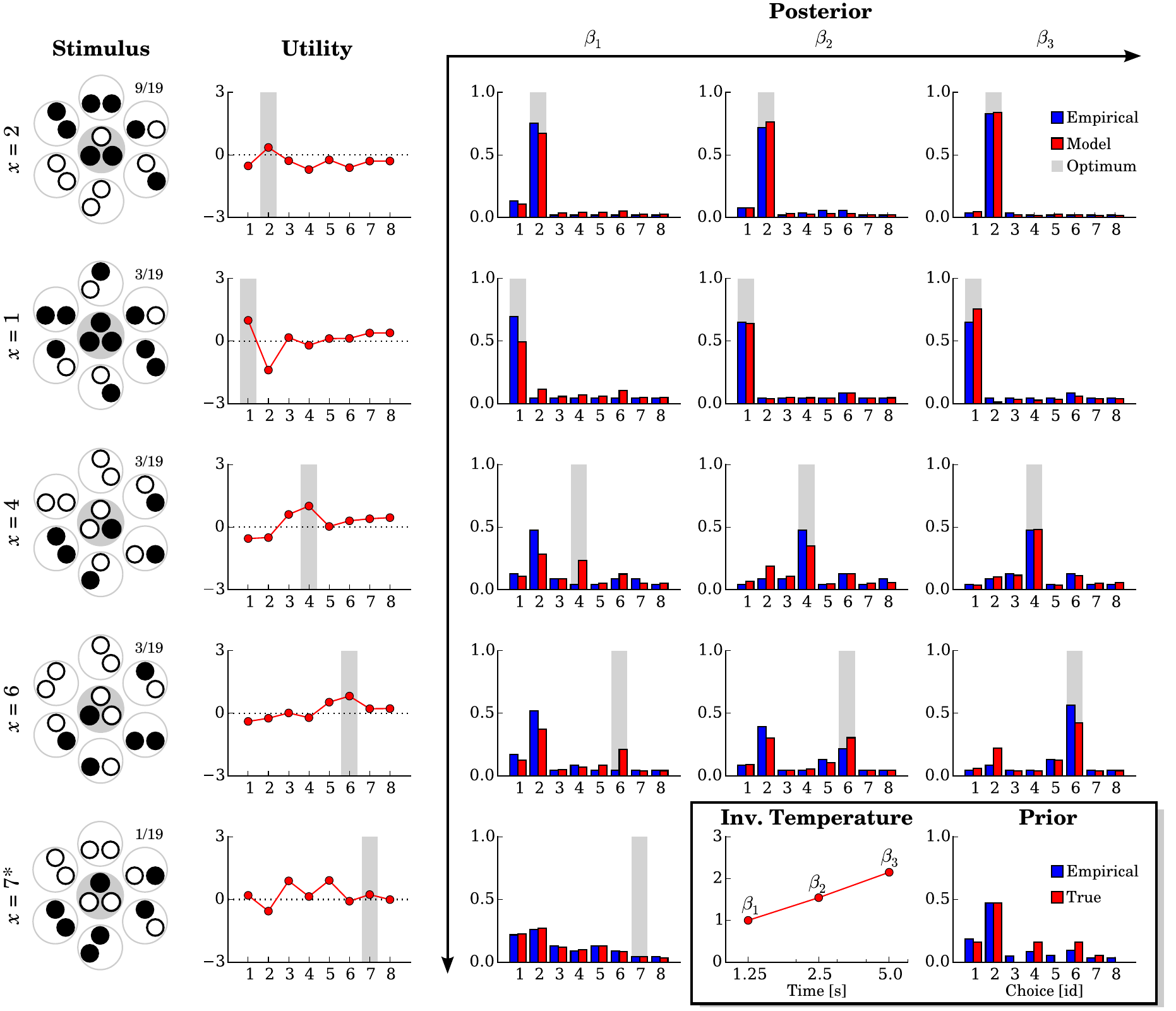}
\par\end{centering}

\caption{\emph{Decomposition of subject 1's posterior choice probabilities}.
Each row corresponds to a different puzzle. The left column shows
each puzzle's stimulus and optimal choice. The posterior distributions
$P(y|x,\beta)$ were decomposed into a prior $P(y)$; a set of time-dependent
inverse temperatures $\beta_{r}$; and a set of stimulus-dependent
utility functions $U_{x}$ over choices, normalized relative to the
certainty-equivalent (\ref{eq:normalize-utilities}). The plots compare
the subject's empirical frequencies against the model fit (in the
\emph{posterior} plots) or against the true optimal choice probabilities
(in the \emph{prior} plot). The stimuli are shown on the left (more
specifically, one out of the 6! arrangement of patches) along with
their probability. Note that the untrained stimulus $x=7$ is the
color-inverse of $x=2$. \label{fig:s1-decomp}}
\end{figure}

For each one of the 30 subjects, we first calculated the empirical
choice probabilities and then estimated their decomposition into an
inverse temperature $\beta$ and utility functions $U_{x}$ using
the procedure detailed in the previous section. The mean error of
the fit was very low ($0.0347\pm0.0024$ bits), implying that the
choice probabilities are well explained by the model. As an example,
Fig.~\ref{fig:s1-decomp} shows the decomposition for subject~1
(error $0.0469$ bits, $83\%$ percentile rank) along with a comparison
between the empirical posterior and the model posterior calculated
from the inferred components using equation~(\ref{eq:model-posterior}).
As durations become longer and $\beta$ increases, the model captures
the gradual shift from the prior towards the optimal choice distribution. 

\begin{figure}
\begin{centering}
\includegraphics[width=1\textwidth]{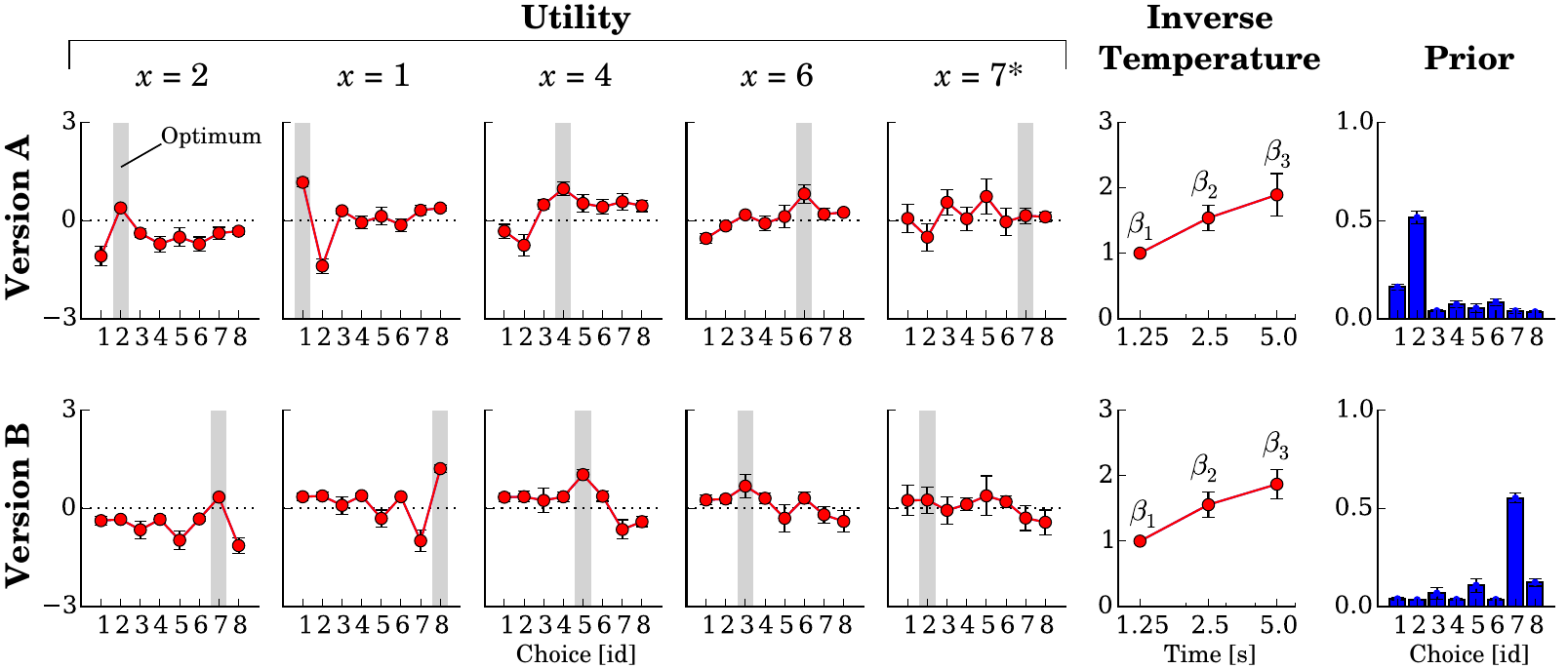}
\par\end{centering}

\caption{\emph{Summary of inferred preferences across all subjects}. The two
rows depict the results for the two versions of the experiment, each
one averaged over 15 subjects. The stimuli of both versions are the
same but with their colors inverted, resulting in a mirror symmetry
along the vertical axis. The figure shows the inferred utility functions
(normalized to the certainty-equivalent); the inverse temperatures;
and the prior over choices. Optimal choices are highlighted in gray.
Error bars denote one standard deviation.\label{fig:all-decomp}}
\end{figure}

As seen in Fig.~\ref{fig:all-decomp}, the resulting decomposition
is stable and shows little variability across subjects. The stimuli
of version~B of the experiment differed from version~A only in that
they were color-inverted, leading to mirror-symmetric decompositions
of the prior and the utility functions. The results suggest the following
trends:
\begin{itemize}
\item \emph{Prior:} Compared to the true distribution over solutions, subjects
tended to concentrate their choices slightly more on the most frequent
optimal solution (i.e.~either $y=2$ or $y=7$ for version~A or~B
respectively) and on the all-black or all-white solution (either $y=1$
or $y=8$). 
\item \emph{Inverse temperature:} The inverse temperature increases monotonically
with longer durations, and the dependency is approximately linear
in log-time (Fig.~\ref{fig:s1-decomp} and~\ref{fig:all-decomp}). 
\item \emph{Utility functions:} In the case of the stimuli that subjects
were trained in (namely, $x\in\{1,2,4,6\}$), the maximum subjective
utility coincides with the solution of the puzzle. Notice that some
choices are enhanced while others are suppressed according to their
subjective utility function. Especially the choice for the most frequent
stimulus ($x=2$) is suppressed when it is suboptimal. In the case
of the untrained stimulus ($x=7$), the utility function is comparatively
flat and variable across subjects.
\end{itemize}
Finally, as a comparison, we also computed the decomposition assuming
a \emph{Softmax function} (or \emph{Boltzmann distribution}):
\begin{equation}
Q(y|x,r)=\frac{1}{Z_{\beta}}\exp\Bigl\{\beta(r)U_{x}(y)\Bigr\}.\label{eq:boltzmann}
\end{equation}
The mean error of the resulting fit was significantly worse (error
$0.0498\pm0.0032$ bits) than the one based on (\ref{eq:model-posterior}),
implying that the inclusion of the prior choice probabilities~$P(y)$
improves the explanation of the choice data.

\subsection{Extrapolation of performance measures}

\begin{figure}
\begin{centering}
\includegraphics[width=0.8\textwidth]{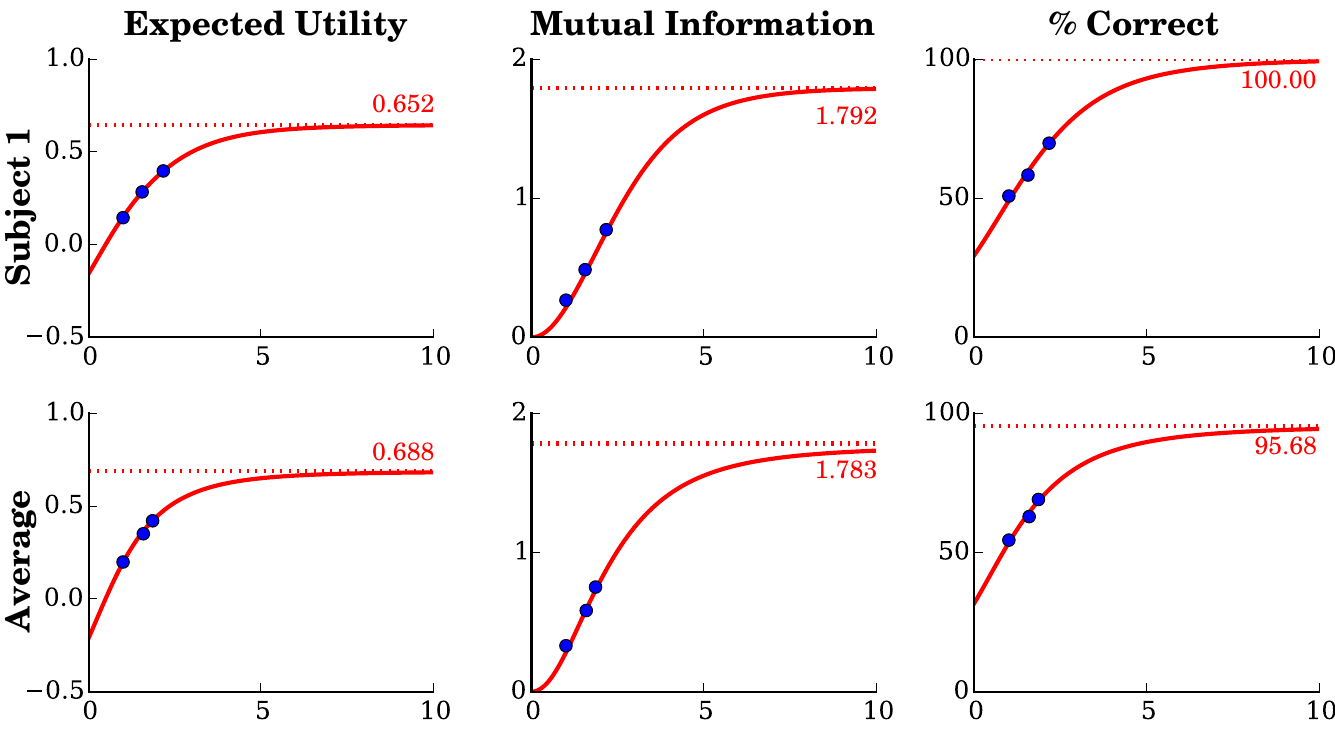}
\par\end{centering}

\caption{\emph{Extrapolation of the performance measures}. The panels show
the expected utility $EU_{\beta}$, the mutual information $I_{\beta}$,
and the expected percentage of correct choices as a function of the
inverse temperature~$\beta$. The top and bottom rows correspond
to subject~1 and the averaged subjects respectively. Each plot shows
the performance measure obtained from the empirical choice probabilities
(blue markers) and the choice probabilities derived from the model
(red curve) together with the maximum attainable value (dotted red).
\label{fig:extrapolation}}
\end{figure}

We calculated the expected utility and the mutual information as a
function of the inverse temperature using~(\ref{eq:eu-i}). The resulting
curves for subject~1 and the average subject are shown in Fig.~\ref{fig:extrapolation}
together with the predicted percentage of correct choices. All the
curves are monotonically increasing and upper bounded. The expected
utility and the percentage of correct choices are concave in the inverse
temperature, indicating marginally diminishing returns with longer
durations. Similarly, the mutual information approaches asymptotically
the upper bound set by the stimulus entropy $H(X)\approx1.792$ bits
(excluding the untrained stimulus). 

\section{Discussion and Conclusion}

It has long been recognized that the model of perfect rationality
does not adequately capture human decision-making because it neglects
the numerous resource limitations that prevent the selection of the
optimal choice \citep{Simon1984}. In this work, we considered a model
of bounded-rational decision-making inspired by ideas from statistical
mechanics and information-theory. A distinctive feature of this model
is the interplay between the decision-maker's preferences, a prior
distribution over choices, and a resource parameter. To test the model,
we conducted an experiment in which participants had to solve puzzles
under time pressure. The experimental results are very well predicted
by the model, which allows us to draw the following conclusions:
\begin{enumerate}
\item \emph{Prior}: When the decision-making resources decrease, people's
choices fall back on a prior distribution. This conclusion is supported
by two observations. First, the bounded-rational model explains the
gradual shift of the subjects' choice probabilities towards the prior
as the duration of the trial is reduced (e.g.~Fig.\ref{fig:s1-decomp}).
Second, the model fit obtained by the Softmax rule~(\ref{eq:boltzmann}),
which differs from the bounded rational model~(\ref{eq:model-posterior})
only by the lack of a prior distribution, has a significantly larger
error. Thus, our results conflict with the predictions made by models
that lack a prior choice distribution---most notably with expected
utility theory \citep{Neumann1944,Savage1954} and the choice models
based on the Softmax function (typical in reinforcement learning,
but also in e.g.~the \emph{logit rule} of \emph{quantal response
equilibria}~\citep{Friedman2005} or in\emph{ maximum entropy inverse
reinforcement learning} \citep{Ziebart2008}).
\item \emph{Utility and Inverse Temperature}: Posterior choice probabilities
can be meaningfully parameterized in terms of utilities (which capture
the decision-maker's preferences) and inverse temperatures (which
encode resource constraints). This is evidenced by the quality of
the fit and the cogent operational role of the parameters. Utilities
are stimulus-contingent enhancers/inhibitors that act upon the prior
choice probabilities, consistent with the role of utility as a measure
of relative desirability in \emph{regret theory} \citep{Bleichrodt2015}
and also related to the cognitive functions attributed to the \emph{dorsal
anterior cingulate cortex} \citep{Shenhav2013}. On the other hand,
the inverse temperature captures a determinant factor of choice behavior
that is independent of the preferences---mathematically embodied in
the low-rank assumption of the log-likelihood function that we used
for the decomposition in the analysis. This assumption does not comply
with the necessary conditions for rational meta-reasoning, wherein
decision-makers can utilize the knowledge about their own resources
in their strategy~\citep{Lieder2014}. 
\item \emph{Preference Learning}: Utilities are learned from experience.
As is seen in the utility functions of Fig.~\ref{fig:all-decomp},
subjects did not learn the optimal choice of the untrained stimulus
(i.e.~$x=7$) in spite of being just a simple color-inversion of
the most frequent stimulus (i.e.~$x=2$). Our experiment did not
address the mechanisms that underlie the acquisition of preferences.
However, given that the information necessary to establish a link
between the stimulus and the optimal choice is below two bits (that
is, far below the ${3 \choose 2}\cdot2^{2}\cdot6=72$ bits necessary
to represent an arbitrary member of the considered class of puzzles),
it is likely that the training phase had subjects synthesize perceptual
features that allowed them to efficiently identify the optimal solution.
Other avenues are explored in \citep{Srivastava2012,Srivastava2014}
and references therein.
\item \emph{Diminishing returns}: The decision-maker's performance is marginally
diminishing in the amount of resources. This is seen in the concavity
of the expected utility curve (Fig.~\ref{fig:extrapolation}; similarly
in the percentage of correct choices) combined with the sub-linear
growth of the inverse temperature as a function of the duration (Fig.~\ref{fig:all-decomp}).
For most subjects, the model predicts a perfectly-rational choice
behavior in the limit of unbounded trial duration.
\end{enumerate}
In summary, in this work we have shown empirically that the model
of bounded rationality provides an adequate explanatory framework
for resource-constrained decision-making in humans. Using a challenging
cognitive task in which we could control the time available to arrive
at a choice, we have shown that human decision-making can be explained
in terms of a trade-off between the gains of maximizing subjective
utilities and the losses due to the deviation from a prior choice
distribution.

\subsubsection*{Acknowledgements}

This work was supported by the Office of Naval Research (Grant N000141110744) 
and the University of Pennsylvania.

\newpage{}

{\small{}\bibliographystyle{plainnat}
\bibliography{bibliography}

\begin{thebibliography}{18}
\providecommand{\natexlab}[1]{#1}
\providecommand{\url}[1]{\texttt{#1}}
\expandafter\ifx\csname urlstyle\endcsname\relax
  \providecommand{\doi}[1]{doi: #1}\else
  \providecommand{\doi}{doi: \begingroup \urlstyle{rm}\Url}\fi

\bibitem[Banerjee et~al.(2005)Banerjee, Merugu, Dhillon, and
  Ghosh]{Banerjee2005}
A.~Banerjee, S.~Merugu, I.~S. Dhillon, and J.~Ghosh.
\newblock {Clustering with Bregman Divergences}.
\newblock \emph{Journal of Machine Learning Research}, 6:\penalty0 1705--1749,
  2005.

\bibitem[Bell(1982)]{Bell1982}
D.E. Bell.
\newblock {Regret in decision making under uncertainty}.
\newblock \emph{Operations Research}, 33:\penalty0 961--981, 1982.

\bibitem[Bleichrodt and Wakker(2015)]{Bleichrodt2015}
H.~Bleichrodt and P.~P. Wakker.
\newblock {Regret theory: A bold alternative to the alternatives}.
\newblock \emph{The Economic Journal}, 125\penalty0 (583):\penalty0 493--532,
  2015.

\bibitem[Fishburn(1982)]{Fishburn1982}
{P.C.} Fishburn.
\newblock \emph{{The Foundations of Expected Utility}}.
\newblock D. Reidel Publishing, Dordrecht, 1982.

\bibitem[Friedman and Mezzetti(2005)]{Friedman2005}
J.W. Friedman and C.~Mezzetti.
\newblock {Random belief equilibrium in normal form games}.
\newblock \emph{Games and Economic Behavior}, 51\penalty0 (2):\penalty0
  296--323, 2005.

\bibitem[Gigerenzer and Selten(2001)]{Gigerenzer2001}
G.~Gigerenzer and R.~Selten.
\newblock \emph{{Bounded rationality: the adaptive toolbox}}.
\newblock {MIT} Press, Cambridge, MA, 2001.

\bibitem[Lieder et~al.(2014)Lieder, Plunkett, Hamrick, Russell, Hay, and
  Griffiths]{Lieder2014}
F.~Lieder, D.~Plunkett, J.~B. Hamrick, S.~J. Russell, N.~Hay, and T.~Griffiths.
\newblock {Algorithm selection by rational metareasoning as a model of human
  strategy selection}.
\newblock \emph{Advances in Neural Information Processing Systems}, pages
  2870--2878, 2014.

\bibitem[Loomes and Sugden(1982)]{Loomes1982}
G.~Loomes and R.~Sugden.
\newblock {Regret theory: {A}n alternative approach to rational choice under
  uncertainty}.
\newblock \emph{Economic Journal}, 92:\penalty0 805--824, 1982.

\bibitem[Ortega and Braun(2013)]{Ortega2013}
P.~A. Ortega and D.~A. Braun.
\newblock {Thermodynamics as a theory of decision-making with
  information-processing costs}.
\newblock \emph{Proceedings of the Royal Society A: Mathematical, Physical and
  Engineering Science}, 469\penalty0 (2153), 2013.

\bibitem[Rubinstein(1998)]{Rubinstein1998}
A.~Rubinstein.
\newblock \emph{{Modeling bounded rationality}}.
\newblock MIT Press, 1998.

\bibitem[Savage(1954)]{Savage1954}
L.J. Savage.
\newblock \emph{{The Foundations of Statistics}}.
\newblock John Wiley and Sons, New York, 1954.

\bibitem[Shenhav et~al.(2013)Shenhav, Botvinick, and Cohen]{Shenhav2013}
A.~Shenhav, M.~M. Botvinick, and J.~D. Cohen.
\newblock {The expected value of control: an integrative theory of anterior
  cingulate cortex function}.
\newblock \emph{Neuron}, 79:\penalty0 217--240., 2013.

\bibitem[Simon(1984)]{Simon1984}
H.~Simon.
\newblock \emph{{Models of Bounded Rationality}}.
\newblock {MIT} Press, Cambridge, MA, 1984.

\bibitem[Srivastava and Schrater(2012)]{Srivastava2012}
N.~Srivastava and P.~R. Schrater.
\newblock {Rational inference of relative preferences}.
\newblock \emph{Advances in neural information processing systems}, 2012.

\bibitem[Srivastava et~al.(2014)Srivastava, Vul, and Schrater]{Srivastava2014}
N.~Srivastava, E.~Vul, and P.~R. Schrater.
\newblock {Magnitude-sensitive preference formation}.
\newblock \emph{Advances in neural information processing systems}, 2014.

\bibitem[Tishby and Polani(2011)]{Tishby2011}
N.~Tishby and D.~Polani.
\newblock {Information Theory of Decisions and Actions}.
\newblock In Hussain~Taylor Vassilis, editor, \emph{{Perception-reason-action
  cycle: Models, algorithms and systems}}. Springer, Berlin, 2011.

\bibitem[{Von Neumann} and Morgenstern(1944)]{Neumann1944}
J.~{Von Neumann} and O.~Morgenstern.
\newblock \emph{{Theory of Games and Economic Behavior}}.
\newblock Princeton University Press, Princeton, 1944.

\bibitem[Ziebart et~al.(2008)Ziebart, Maas, Bagnell, and Dey]{Ziebart2008}
B.~D. Ziebart, A.~L. Maas, J.~A. Bagnell, and A.~K. Dey.
\newblock {Maximum Entropy Inverse Reinforcement Learning}.
\newblock In \emph{{AAAI}}, pages 1433--1438, 2008.

\end{thebibliography}
}{\small \par}

\section*{}
\end{document}